\documentclass[letterpaper, 10 pt, conference]{ieeeconf}

\IEEEoverridecommandlockouts     
\usepackage{graphicx}
\usepackage{svg}                 
\usepackage{amsmath,amssymb,amsfonts}
\usepackage{subcaption}          
\usepackage{booktabs}            
\usepackage{xcolor}
\usepackage{tikz}
\usetikzlibrary{calc,shapes.geometric}

\newtheorem{theorem}{Theorem}
\newtheorem{lemma}{Lemma}
\newtheorem{proposition}{Proposition}

\newtheorem{definition}{Definition}
\newtheorem{assumption}{Assumption}
\newtheorem{problem}{Problem}
\newtheorem{remark}{Remark}
\makeatletter
\renewenvironment{proof}%
  {\par\noindent\hspace{2em}{\itshape Proof: }}%
  {\hspace*{\fill}~\QED\par\addvspace{0.5\baselineskip}}
\makeatother

\newcommand{\R}{\mathbb{R}}

\title{\LARGE \bf
Nested Field-of-View Barriers for \\ Teleoperated Multi-Robot Relay Chains
}

\author{Arthur Vangasse$^{1}$ Theo Cruz$^{1}$ Thales Soares$^{1}$ Andr\'e Cid$^{1}$ \\ Vin\'icius Gon\c{c}alves$^{2}$ Gustavo Freitas$^{2}$ Luciano Pimenta$^{2}$
\thanks{*This work was partially supported by CNPq, FAPEMIG, CAPES, and Petrobras/ANP under grant 2023/00494-5 and also grant under contract number 0050.0128906.24.9.}%
\thanks{$^{1}$The authors are with the Graduate Program in Electrical Engineering,
        Federal University of Minas Gerais, Belo Horizonte, Brazil
        {\tt\small \{vangasse, theofonsecacruz, thalesasoares, andrecid\}@ufmg.br}}%
\thanks{$^{2}$The authors are with the School of Engineering, Federal University of Minas Gerais,
        Belo Horizonte, Brazil
        {\tt\small \{gustavomfreitas, marianoauto, lucpim\}@ufmg.br}}%
}

\begin{document}

\maketitle
\thispagestyle{empty}
\pagestyle{empty}

\begin{abstract}
Teleoperation of mobile robots in communication-denied environments requires maintaining a communication link while allowing the operator to maneuver freely. We propose a shared-autonomy framework in which a teleoperated leader is followed by a chain of autonomous robots that relays a line of sight to a fixed base station, a geometric proxy for the link. Connectivity is enforced geometrically through visibility-based Control Barrier Functions (CBFs), with each agent computing its control input locally through a quadratic program. A strict visibility constraint guarantees connectivity, but a follower with no independent motion objective remains stationary, stalling the leader at the boundary of the safe set. Relaxing it preserves mobility at the cost of the guarantee. Our framework resolves this conflict through nested hard-soft visibility constraints: a full field-of-view (FoV) constraint enforced strictly to guarantee connectivity and a contracted FoV constraint imposed on each follower as a relaxed behavioral objective that drives it inward before the connectivity boundary is reached. We prove feasibility of the local quadratic programs and forward invariance of the connected, collision-free set under the resulting distributed controller. The contraction parameter shapes follower behavior without affecting the guarantee, which follows from the strict constraint alone. The framework is evaluated in confined and cluttered environments on two simulated systems: a kinematic integrator and software-in-the-loop \textit{Crazyflie} quadrotors. Compared with a baseline tuned to reach the goal in comparable time, the framework keeps the links far closer to the connected set and moves the followers less.
\end{abstract}

\section{Introduction}
\label{sec:introduction}

\begin{figure}[t]
\centering
\begin{tikzpicture}[scale=1.09,>=stealth,font=\footnotesize,
  fov/.style={fill=green!55!black,fill opacity=0.11},
  fovb/.style={draw=green!45!black,line width=0.4pt,line join=round},
  fovn/.style={draw=green!45!black,densely dotted,line width=0.4pt},
  shad/.style={fill=gray!55,fill opacity=0.16},
  sight/.style={draw=black!55,densely dotted,line width=0.4pt,
                shorten >=6.5pt,shorten <=6.5pt},
  cf/.style={draw=green!35!black,densely dashed,line width=0.5pt},
  rock/.style={fill=gray!55,draw=gray!75},
  rel/.style={draw=black,fill=orange!75},
  led/.style={draw=black,fill=blue!60},
  bse/.style={draw=black,fill=black!65}]

\coordinate (B)  at (0,0);
\coordinate (A3) at (2.1,0.3);
\coordinate (A2) at (4.15,-0.2);
\coordinate (A1) at (6.1,0.35);

\fill[fov]  (B)  -- ++(-17.9:2.5) arc (-17.9:34.1:2.5)  -- cycle;
\draw[fovb] (B)  -- ++(-17.9:2.5) arc (-17.9:34.1:2.5)  -- cycle;
\fill[fov]  (A3) -- ++(-39.7:2.5) arc (-39.7:12.3:2.5)  -- cycle;
\draw[fovb] (A3) -- ++(-39.7:2.5) arc (-39.7:12.3:2.5)  -- cycle;

\def\fovthree{(A2) -- ++(-10.3:2.5) arc (-10.3:41.7:2.5) -- cycle}
\def\occluder{(5.4655,0.3640) arc (-66.79:-214.09:0.42)
              -- (5.9995,2.5330) -- (7.1830,1.1003) -- cycle}
\begin{scope}
  \clip \fovthree;
  \fill[shad] \occluder;
  \draw[fovn] \fovthree;
\end{scope}
\begin{scope}[even odd rule]
  \clip (-0.7,-1.7) rectangle (8.3,2.7) \occluder;
  \fill[fov]  \fovthree;
  \draw[fovb] \fovthree;
\end{scope}
\fill[rock] (5.3,0.75) circle (0.42);
\node at (5.3,0.75) {$\mathcal{O}$};
\begin{scope}
  \clip \fovthree;
  \draw[fovb] \occluder;
\end{scope}

\draw[cf] (3.097,0.057) -- (3.926,-0.631) arc (-27.02:-0.38:2.05) -- cycle;
\draw[<->,green!35!black] (3.956,0.705) -- (4.052,0.265);
\node[green!30!black,anchor=west,inner sep=1.5pt] at (4.10,0.50) {$\epsilon\bar{d}$};

\node[green!30!black,inner sep=1pt] at (1.05,0.50) {$\bar{\mathcal{F}}$};
\node[green!30!black,inner sep=1pt] at (5.70,-0.12) {$\mathcal{F}$};
\draw[cf] (3.82,-0.71) -- (3.87,-0.56);
\node[green!30!black,inner sep=1pt] at (3.80,-0.82) {$\mathcal{F}^c$};
\draw[fovb] (0.78,0.96) -- (1.02,0.70);
\node[green!30!black,anchor=east,inner sep=1.5pt] at (0.76,0.98) {$\partial\mathcal{F}$};
\draw[<->] (A3) -- ++(-39.7:2.5);
\node[anchor=north,inner sep=2pt] at (3.06,-0.52) {$d_{\max}$};

\draw[sight,->] (B)  -- (A3);
\draw[sight,->] (A3) -- (A2);
\draw[sight,->] (A2) -- (A1);

\draw[->,very thick,blue!60!black] (A1) -- ++(0.75,0.28);

\draw[bse] ($(B)+(-0.18,-0.18)$) rectangle ($(B)+(0.18,0.18)$);
\draw[rel] (A3) circle (0.2);
\draw[rel] (A2) circle (0.2);
\draw[led] (A1) circle (0.2);

\draw[bse] (-0.06,-0.97) rectangle (0.16,-0.75);
\node[anchor=west,inner sep=2pt] at (0.22,-0.86) {base};
\draw[rel] (0.05,-1.08) circle (0.11);
\node[anchor=west,inner sep=2pt] at (0.22,-1.08) {relays};
\draw[led] (0.05,-1.30) circle (0.11);
\node[anchor=west,inner sep=2pt] at (0.22,-1.30) {leader};
\node[blue!50!black,anchor=west,inner sep=1.5pt] at (6.9,0.67) {$\mathbf{r}_1$};
\end{tikzpicture}
\caption{The nested field-of-view framework. A teleoperated leader, a chain of relays each holding its predecessor inside a field of view $\mathcal{F}$ that obstacles cut, and a fixed base.}
\label{fig:chain}
\end{figure}
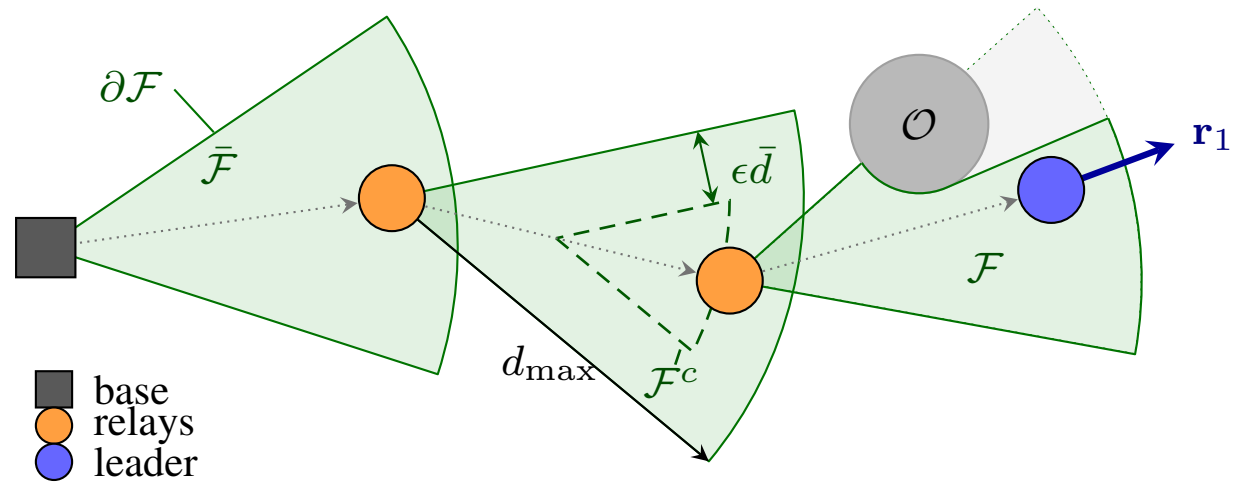

Shared-autonomy teleoperation blends the intent of a human operator with the reactive safety of onboard autonomy \cite{li2023the}, but it works only for as long as the communication link holds. In cluttered or subterranean workspaces, non-line-of-sight propagation and severe attenuation \cite{rappaport1996wireless, cid2024wireless} separate the operator from the robot through delay and dropout \cite{khasawneh2019human, gulecyuz2025enhancing} long before its endurance is spent. We propose a framework in which the robots themselves become the infrastructure. A leader is teleoperated, while a chain of autonomous followers positions itself so that line-of-sight (LoS) is relayed from the leader back to a fixed base station (Fig.~\ref{fig:chain}). Connectivity is treated geometrically, and LoS is seen as a proxy for the link rather than a model of it. Each follower keeps its predecessor inside a field of view (FoV) that is bounded in range, limited in aperture, and cut by the obstacles within it; the shaded wedge of Fig.~\ref{fig:chain} is the part of a FoV that an obstacle denies.

In the DARPA Subterranean Challenge, leading teams extended their range by dropping static radio relays as they advanced \cite{ohradzansky2022multi, tranzatto2022team, kottege2025heterogeneous}. Such nodes are simple and effective, but committed irrevocably at drop time and left behind; we evolve from that solution by keeping the relays mobile. Other works deploy mobile robots as relays using signal-strength heuristics \cite{Laclau2021} or alternate between exploration and chain modes \cite{Tian2025}, while network-aware teleoperation adapts to link quality but presumes that an infrastructure is already in place \cite{chen2024network}. We instead make connectivity a deterministic geometric constraint, imposed by Control Barrier Functions (CBFs) \cite{ames2019control, ames2017control}, which have been applied to connectivity maintenance \cite{lee2025maintaining, ong2023nonsmooth, bhatia2024decentralized, capelli2020connectivity} and LoS preservation \cite{pan2025robusttrajectorygenerationcontrol, BaiICRA2025}.

Our visibility constraint is adapted from the FoV barriers developed for pursuit-evasion under occlusion \cite{koga2023hide, zhou2025control}, whose scope differs decisively from ours. There, keeping the target in view is the pursuer's own objective, and the barrier certifies a motion that the nominal controller already produces. Here, visibility is a supporting requirement, not an objective. Followers carry no reference of their own and move only when a constraint compels them.

That absence of a reference is what makes the constraint hard to use. The hard barrier is minimally invasive by construction, reacting only once the FoV boundary is met. The team therefore rides that boundary. The leader is braked at the limit, each follower advances no further than it is forced to, and the chain stalls at the border of the safe set. Slack variables restore mobility but surrender the guarantees. Inspired by the approach in \cite{roncero2025multi}, we combine two nested CBFs on each neighboring pair. The full FoV is enforced strictly on both agents so that the connected set stays forward invariant. A contracted copy of it, set in from the range limit and from the aperture edges by a margin $\epsilon\bar{d}$ and drawn dashed in Fig.~\ref{fig:chain}, is imposed on the follower alone as a relaxed constraint. It supplies the drive the follower lacks and pulls it inward before the strict boundary is reached, yielding whenever the environment makes that unattainable. We nonetheless keep that minimal invasiveness. Giving the followers a reference of their own would resolve the stall by turning the framework into a formation controller, committing each relay to a shape fixed in advance; here a relay with no objective is moved by the relaxed constraint alone, while one carrying a task of its own has that task filtered rather than replaced.

We contribute (i) a nested hard-soft formulation that separates the connectivity guarantee from the behavior that sustains the chain, (ii) a distributed synthesis with proofs of feasibility and forward invariance of the connected, collision-free set, and (iii) an evaluation on two simulated systems and two workspaces that traces the residual violations to tracking error and to field-of-view resolution.

\section{Preliminaries}
\label{sec:preliminaries}

\subsection{Agent Model}
\label{sec:agent_model}

Each agent is modeled as a single integrator,
\begin{equation}
    \dot{\mathbf{x}}(t) = \mathbf{u}(t),
    \label{eq:integrator}
\end{equation}
where $\mathbf{x} \in \R^{n}$ is the state and $\mathbf{u} \in \mathcal{U} \subset \R^{n}$ is the velocity command, with $\mathcal{U}$ compact, convex, and containing $\mathbf{0}$.

The abstraction places the safety layer at the velocity-control level, assuming the onboard controller tracks the command within a control period.  The visibility and collision barriers then have relative degree one and enter the quadratic program as constraints affine in the command.

\subsection{Control Barrier Functions}
\label{sec:cbf}

Safety is expressed as the requirement that the state remains within a set $\mathcal{S}$, defined as the superlevel set of a continuously differentiable function $h: \R^n \to \R$,
\begin{equation}
    \mathcal{S} = \{ \mathbf{x} \in \R^n \mid h(\mathbf{x}) \ge 0 \}.
    \label{eq:safe_set}
\end{equation}

\begin{definition}[\cite{ames2019control}]
\label{def:cbf}
A continuously differentiable function $h: \R^n \to \R$ is a CBF for \eqref{eq:integrator} on a set $\mathcal{D} \supseteq \mathcal{S}$ if there exists an extended class $\mathcal{K}$ function $\alpha: \R \to \R$ such that
\begin{equation}
    \sup_{\mathbf{u} \in \mathcal{U}} \left\{ \frac{\partial h}{\partial \mathbf{x}}(\mathbf{x})\, \mathbf{u} \right\} \ge -\alpha(h(\mathbf{x})), \quad \forall \mathbf{x} \in \mathcal{D}.
\end{equation}
\end{definition}

Definition~\ref{def:cbf} asks only that such an input exist; on the boundary $\partial\mathcal{S}$, where $h = 0$, it forbids $h$ from decreasing at all. Choosing such an input at every state keeps the boundary from being crossed, as the following result states. 

\begin{lemma}[\cite{ames2019control}]
\label{lem:invariance}
Let $h$ be a CBF for \eqref{eq:integrator}. Any locally Lipschitz continuous controller $\mathbf{u}(\mathbf{x})$ satisfying
\begin{equation}
    \frac{\partial h}{\partial \mathbf{x}}(\mathbf{x})\, \mathbf{u} \ge -\alpha(h(\mathbf{x})), \quad \forall \mathbf{x} \in \mathcal{S},
    \label{eq:cbf_condition}
\end{equation}
renders $\mathcal{S}$ forward invariant.
\end{lemma}

\subsection{Signed Distance Function}
\label{sec:sdf}

Remaining outside an obstacle or inside a field of view is a condition of membership in a set, which a barrier function \eqref{eq:safe_set} can express through any scalar whose sign encodes that membership. We adopt the signed distance introduced in the following definition.

\begin{definition}
\label{def:sdf}
The signed distance function $d: \R^2 \times 2^{\R^2} \to \R$ from a point $\mathbf{p}$ to a closed set $\mathcal{F} \subset \R^2$ with boundary $\partial\mathcal{F}$ is
\begin{equation}
    d(\mathbf{p}, \mathcal{F}) =
    \begin{cases}
        -\min_{\mathbf{p}^* \in \partial\mathcal{F}} \|\mathbf{p} - \mathbf{p}^*\|, & \mathbf{p} \in \mathcal{F}, \\
        \ \ \ \min_{\mathbf{p}^* \in \partial\mathcal{F}} \|\mathbf{p} - \mathbf{p}^*\|, & \mathbf{p} \notin \mathcal{F}.
    \end{cases}
    \label{eq:sdf}
\end{equation}
\end{definition}

The function is negative inside the set, positive outside, and zero on its boundary, so that $\mathbf{p} \in \mathcal{F}$ if and only if $d(\mathbf{p}, \mathcal{F}) \le 0$. For a fixed set it is $1$-Lipschitz in $\mathbf{p}$ \cite{zhou2025control}, and where it is differentiable its gradient is the unit vector pointing away from the closest boundary point.

\subsection{Control Synthesis via Quadratic Programming}
\label{sec:qp}

The constraint \eqref{eq:cbf_condition} is affine in $\mathbf{u}$ for the integrator \eqref{eq:integrator}, so a controller that satisfies a collection of such constraints while deviating minimally from a reference command $\mathbf{r}$ is obtained from a strictly convex quadratic program (QP) \cite{ames2017control},
\begin{equation}
\begin{aligned}
    \mathbf{u}^*(\mathbf{x}) = \arg\min_{\mathbf{u} \in \mathcal{U}} \ & (\mathbf{u} - \mathbf{r})^\top \mathbf{P} (\mathbf{u} - \mathbf{r}) \\
    \text{s.t.} \ \ & \frac{\partial h_s}{\partial \mathbf{x}}(\mathbf{x})\, \mathbf{u} \ge -\alpha_s\, h_s(\mathbf{x}), \quad s \in \mathcal{C},
\end{aligned}
\label{eq:cbf_qp}
\end{equation}
where $\mathbf{P} \succ 0$ weights the deviation, $\mathcal{C}$ indexes the barriers $h_s$ imposed on the agent, and the constant  $\alpha_s > 0$, a linear choice of the extended class $\mathcal{K}$ function in \eqref{eq:cbf_condition}, sets how aggressively each boundary may be approached. Strictly convex QPs are solved efficiently in real time and, when feasible, admit a unique solution \cite{wright2015continuous}. For our formulation, two properties of \eqref{eq:cbf_qp} matter: its feasibility and the role of slack variables. The following remark addresses both.

\begin{remark}
\label{rem:feasibility}
For \eqref{eq:integrator}, the input $\mathbf{u} = \mathbf{0} \in \mathcal{U}$ satisfies every constraint of \eqref{eq:cbf_qp} wherever the barriers are nonnegative, so the program is feasible there without recourse to relaxation. A constraint may still be relaxed by a slack $\delta \ge 0$ penalized in the cost, in the form $\frac{\partial h_s}{\partial \mathbf{x}}\, \mathbf{u} \ge -\alpha_s h_s - \delta$, but here a slack shapes behavior rather than restores feasibility, unlike \cite{zhou2025control}, where an exogenous evader makes the visibility barrier time-varying and the slack is needed to keep the program feasible.
\end{remark}

\section{Problem Formulation}
\label{sec:problem}

A team of $N$ agents operates in a planar workspace, indexed by $\mathcal{N} = \{1, \dots, N\}$. The pose of agent $i$ is $\mathbf{x}_i = [\mathbf{p}_i^\top, \theta_i]^\top \in \R^3$, collecting its position $\mathbf{p}_i \in \R^2$ and heading $\theta_i \in \R$, the latter taken as the unwrapped angle rather than an element of $S^1$ so that the state remains in $\R^n$, as Section~\ref{sec:preliminaries} assumes. The team state $\mathbf{x} \in \R^{3N}$ stacks the $\mathbf{x}_i$ vertically, and each agent evolves under \eqref{eq:integrator} as $\dot{\mathbf{x}}_i = \mathbf{u}_i$.

Agent $1$ is the leader and the only agent to receive a reference, $\mathbf{r}_1(t) \not\equiv \mathbf{0}$, a desired velocity supplied by the operator through an input device such as a joystick. Agent $N$ is a fixed base station, $\mathbf{u}_N \equiv \mathbf{0}$. The remaining agents carry no reference of their own, $\mathbf{r}_i \equiv \mathbf{0}$ for $1 < i < N$, and move only when a constraint compels them. Consecutive agents form the relay pairs indexed by $\mathcal{E} = \{2, \dots, N\}$. Within pair $i$, agent $i-1$ is \emph{observed} and agent $i$ \emph{observes}.

Communication within a pair is modeled geometrically. Each agent carries a sensor whose FoV is bounded in range and aperture and cut by the obstacles $\mathcal{O}$, and pair $i$ is connected when the observed agent lies in the FoV of the observer. The team is connected to the base when every pair is, and safe when no agent collides with another or with an obstacle. Both properties hold initially and must be preserved while the operator drives the leader.

\begin{problem}
\label{prob:main}
Design a distributed control law under which each agent computes its own velocity command from its state, the states of its chain neighbors, the positions of the agents in its vicinity, and the geometry of the obstacles $\mathcal{O}$, such that (i) every pair remains connected and every agent collision-free for all time, (ii) the leader tracks $\mathbf{r}_1(t)$ as closely as (i) permits, and (iii) the followers sustain the chain without a reference of their own.
\end{problem}

\section{Methodology}
\label{sec:methodology}

The requirements of Problem~\ref{prob:main} are met within the CBF framework of Section~\ref{sec:preliminaries}. Connectivity and collision avoidance are expressed as barrier functions built from signed distances, and each agent computes its velocity command from a local program of the form \eqref{eq:cbf_qp}. Under this program, the leader tracks the operator command until a constraint becomes active, while the followers move only as the constraints require. Two visibility constraints act on each relay pair: a strict one on the full FoV, which carries the connectivity guarantee, and a relaxed one on a contracted FoV, which sets the followers into motion. This section defines these barriers and the safe set they delimit, together with the assumptions under which the distributed law is analyzed in the sequel.

\subsection{Visibility and the Contracted Field of View}
\label{sec:two_fovs}

Let $\bar{\mathcal{F}}(\mathbf{x}_i) \subset \R^2$ denote the nominal FoV of agent $i$, bounded in range and aperture, and let $\mathcal{F}(\mathbf{x}_i) \subseteq \bar{\mathcal{F}}(\mathbf{x}_i)$ be its restriction under occlusion by the obstacles $\mathcal{O}$ (Fig.~\ref{fig:chain}). Both depend on $\mathbf{x}_i$ alone. Write $d_{\max}$ for the range of the nominal FoV, the greatest distance at which the observer can see. Write $\bar{d} = -\min_{\mathbf{p}} d(\mathbf{p}, \bar{\mathcal{F}})$ for the depth of the nominal FoV, the radius of the largest ball it contains, which is constant because $\bar{\mathcal{F}}$ moves rigidly with the agent. The contraction of Definition~\ref{def:contracted} is the fraction $\epsilon$ of that depth, so $\epsilon < 1$ leaves the contracted nominal FoV nonempty.

Pair $i$ is connected when the observed agent lies in the observer's occluded FoV, $\mathbf{p}_{i-1} \in \mathcal{F}(\mathbf{x}_i)$, which by Definition~\ref{def:sdf} is the condition $d(\mathbf{p}_{i-1}, \mathcal{F}(\mathbf{x}_i)) \le 0$. The corresponding barrier is
\begin{equation}
    h_i(\mathbf{x}) = -\, d\big(\mathbf{p}_{i-1}, \mathcal{F}(\mathbf{x}_i)\big), \qquad i \in \mathcal{E}.
    \label{eq:h_full}
\end{equation}

The second, tighter FoV is obtained by eroding the first.

\begin{definition}
\label{def:contracted}
For a contraction parameter $\epsilon \in (0,1)$, the \emph{contracted FoV} of agent $i$ is the erosion of $\mathcal{F}(\mathbf{x}_i)$ by a closed ball of radius $\epsilon \bar{d}$,
\begin{equation}
    \mathcal{F}^c(\mathbf{x}_i) = \big\{ \mathbf{p} \in \R^2 \mid d(\mathbf{p}, \mathcal{F}(\mathbf{x}_i)) \le -\epsilon \bar{d} \big\}.
    \label{eq:contracted_fov}
\end{equation}
\end{definition}

Because \eqref{eq:contracted_fov} is a sublevel set of the same signed distance that defines \eqref{eq:h_full}, the contracted FoV is described by a barrier differing from $h_i$ by a constant,
\begin{equation}
    g_i(\mathbf{x}) = h_i(\mathbf{x}) - \epsilon \bar{d}, \qquad i \in \mathcal{E},
    \label{eq:h_contracted}
\end{equation}
so that $\mathbf{p}_{i-1} \in \mathcal{F}^c(\mathbf{x}_i)$ if and only if $g_i(\mathbf{x}) \ge 0$. The two barriers are nested, $\{ g_i \ge 0 \} \subset \{ h_i \ge 0 \}$ strictly, and share a gradient, $\partial g_i / \partial \mathbf{x} = \partial h_i / \partial \mathbf{x}$.
\subsection{Collision Avoidance}
\label{sec:safety_sets}

Agents are modeled as disks of inflated radius $R$ exceeding their physical radius. For a pair $i \neq j$ and a minimum obstacle clearance $d_o > 0$, the safety barriers are
\begin{equation}
    h_{a,ij}(\mathbf{x}) = \|\mathbf{p}_i - \mathbf{p}_j\| - 2R, \quad
    h_{o,ik}(\mathbf{x}) = \|\mathbf{p}_i - \mathbf{o}_k^*\| - d_o,
    \label{eq:h_safety}
\end{equation}
where $\mathbf{o}_k^*(\mathbf{p}_i) = \arg\min_{\mathbf{o} \in \mathcal{O}_k} \|\mathbf{p}_i - \mathbf{o}\|$ is the point of obstacle $\mathcal{O}_k$ closest to agent $i$. By the Envelope Theorem \cite{simon1994mathematics}, and with the agents circular so that the heading component vanishes,
\begin{equation}
    \frac{\partial h_{o,ik}}{\partial \mathbf{x}_i} = \left[ \frac{(\mathbf{p}_i - \mathbf{o}_k^*)^\top}{\|\mathbf{p}_i - \mathbf{o}_k^*\|} \ \ 0 \right], \quad
    \frac{\partial h_{a,ij}}{\partial \mathbf{x}_i} = \left[ \frac{(\mathbf{p}_i - \mathbf{p}_j)^\top}{\|\mathbf{p}_i - \mathbf{p}_j\|} \ \ 0 \right]
    \label{eq:grad_safety}
\end{equation}
and $\partial h_{a,ij} / \partial \mathbf{x}_j = -\partial h_{a,ij} / \partial \mathbf{x}_i$.

\subsection{Safe Set and Standing Assumptions}
\label{sec:safe_set}

The team is required to remain in the intersection
\begin{equation}
    \mathcal{S} = \mathcal{S}_v \cap \mathcal{S}_a \cap \mathcal{S}_o,
    \label{eq:team_safe_set}
\end{equation}
of the connected set $\mathcal{S}_v = \{ \mathbf{x} \mid h_i(\mathbf{x}) \ge 0, \ \forall i \in \mathcal{E} \}$, the inter-agent set $\mathcal{S}_a = \{ \mathbf{x} \mid h_{a,ij}(\mathbf{x}) \ge 0, \ \forall i \neq j \}$, and the obstacle set $\mathcal{S}_o = \{ \mathbf{x} \mid h_{o,ik}(\mathbf{x}) \ge 0, \ \forall i, \forall \mathcal{O}_k \subset \mathcal{O} \}$.

\begin{remark}
\label{rem:nonsmooth}
The barriers of this paper are built from signed distances, and are not differentiable everywhere. Differentiability fails on the medial axis of the set, where the closest boundary point is not unique, and at its vertices \cite{zhou2025control}. At such points the gradient in \eqref{eq:cbf_condition} is replaced by Clarke's generalized gradient, and the inequality is required to hold for every element of it \cite[Rem.~2, Thm.~3]{glotfelter2017nonsmooth}, a condition that needs only Lipschitz continuity. We impose \eqref{eq:cbf_condition} with a single element of that set, the gradient of the distance to the nearest feature, which is the exact condition wherever the barrier is differentiable. The classical gradient notation is used throughout, with this reading understood at the exceptional points.
\end{remark}

Keeping the team in $\mathcal{S}$ with a distributed law rests on five assumptions: the initial condition, the information available to each agent, the regularity of the visibility barriers, the regularity of the resulting control law, and the regularity of the solutions it generates.

\begin{assumption}
\label{as:init}
The team starts safe and connected, $\mathbf{x}(0) \in \mathcal{S}$.
\end{assumption}

\begin{assumption}
\label{as:info}
Each agent $i$ evaluates its constraints in continuous time using the exact current state of itself and of its chain neighbors $i-1$ and $i+1$, the positions of the agents within its safety radius, and the obstacles $\mathcal{O}$.
\end{assumption}

\begin{assumption}
\label{as:barrier}
Each visibility barrier $h_i$, $i \in \mathcal{E}$, is locally Lipschitz on $\mathcal{S}$.
\end{assumption}

\begin{assumption}
\label{as:lipschitz}
The control law $\mathbf{u}_i(\mathbf{x})$ of every agent $i \in \mathcal{N}$ is locally Lipschitz on $\mathcal{S}$.
\end{assumption}

\begin{assumption}
\label{as:transversal}
Along every closed-loop solution, the set of times at which the team state lies in the nondifferentiability set of a barrier has zero Lebesgue measure.
\end{assumption}

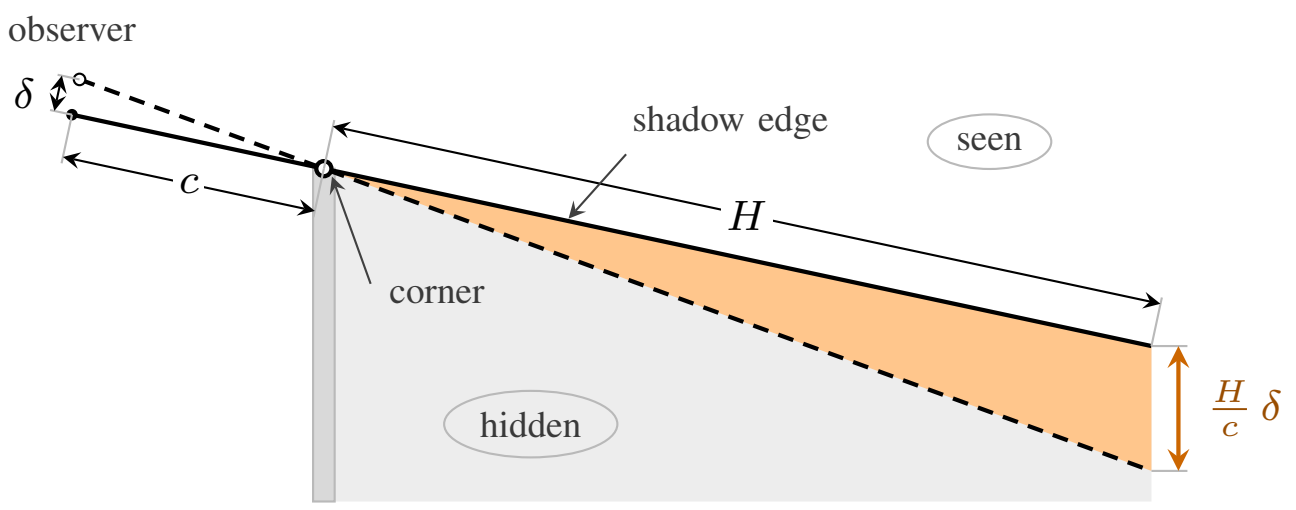
\begin{figure}[t]
\centering
\vspace*{4pt}
\begin{tikzpicture}[scale=1.19,>=stealth,font=\footnotesize,
  wall/.style={fill=gray!30,draw=gray!55},
  hid/.style={fill=black!7},
  sliv/.style={fill=orange!45},
  dim/.style={<->,thin},
  ext/.style={thin,gray!55},
  tag/.style={inner sep=1pt,fill=white},
  key/.style={gray!45!black,font=\scriptsize},
  lead/.style={->,thin,gray!50!black,shorten >=0.5pt}]
  \fill[hid] (0,0) -- (4.6,-0.986) -- (4.6,-1.85) -- (0.06,-1.85) -- cycle;
  \fill[sliv] (0,0) -- (4.6,-0.986) -- (4.6,-1.680) -- cycle;
  \fill[wall] (-0.06,-1.85) rectangle (0.06,0);
  \draw[thick] (-1.4,0.30) -- (4.6,-0.986);
  \draw[thick,densely dashed] (-1.358,0.496) -- (4.6,-1.680);
  \fill (-1.4,0.30) circle (0.032);
  \draw[fill=white] (-1.358,0.496) circle (0.032);
  \draw[fill=white,thick] (0,0) circle (0.042);
  \draw[lead] (0.26,-0.64) -- (0.04,-0.03);
  \node[key,anchor=west,inner sep=2pt] at (0.30,-0.70) {corner};
  \draw[ext] (-1.4,0.30) -- (-1.458,0.030);
  \draw[ext] (0,0) -- (-0.058,-0.270);
  \draw[dim] (-1.450,0.065) -- (-0.050,-0.235);
  \node[tag] at (-0.750,-0.085) {$c$};
  \draw[ext] (0,0) -- (0.058,0.270);
  \draw[ext] (4.6,-0.986) -- (4.658,-0.716);
  \draw[dim] (0.050,0.235) -- (4.650,-0.751);
  \node[tag] at (2.350,-0.258) {$H$};
  \draw[ext] (-1.4,0.30) -- (-1.522,0.326);
  \draw[ext] (-1.358,0.496) -- (-1.480,0.522);
  \draw[dim] (-1.498,0.321) -- (-1.456,0.517);
  \node[anchor=east,inner sep=2pt] at (-1.54,0.42) {$\delta$};
  \draw[ext] (4.6,-0.986) -- (4.80,-0.986);
  \draw[ext] (4.6,-1.680) -- (4.80,-1.680);
  \draw[dim,thick,orange!80!black] (4.75,-0.986) -- (4.75,-1.680);
  \node[anchor=west,orange!60!black,inner sep=2pt] at (4.84,-1.333) {$\tfrac{H}{c}\,\delta$};
  \node[key,anchor=south] at (-1.40,0.60) {observer};
  \node[key,ellipse,draw=gray!55,line width=0.4pt,inner sep=2pt] at (1.15,-1.42) {hidden};
  \node[key,ellipse,draw=gray!55,line width=0.4pt,inner sep=2pt] at (3.70,0.15) {seen};
  \draw[lead] (1.68,0.08) -- (1.35,-0.289);
  \node[key,anchor=south west,inner sep=2pt] at (1.65,0.10) {shadow edge};
\end{tikzpicture}
\caption{A step $\delta$ pivots the shadow edge about the corner, sweeping it by $(H/c)\,\delta$.}
\label{fig:shadow_lever}
\end{figure}

Assumption~\ref{as:barrier} does not follow from the Lipschitz continuity of a distance to a rigidly moving set, because occlusion makes the FoV change shape as the observer moves. Of the pieces bounding that set, the range arc and the aperture edges travel with the observer and the obstacle boundaries do not travel at all, so only the shadow edges (Fig.~\ref{fig:shadow_lever}), which an obstacle corner casts when it cuts the field of view, can move further than the observer itself. Each runs from the observer through the corner that casts it, so the corner is a fixed point of the edge and a step by the observer swings the edge about it, shifting its points $H/c$ times as far as the observer moves, where $H$ is their distance from the corner and $c$ the observer's. On $\mathcal{S}_o$ both are bounded, $H$ by the range $d_{\max}$ and $c$ below by $d_o$ through \eqref{eq:h_safety}, so the rate is at most $d_{\max}/d_o$. 

Assumption~\ref{as:transversal} excludes sustained sliding along a set on which two features of a FoV, or of an obstacle, are equidistant from the point whose distance the barrier measures, and holds whenever such sets are crossed transversally. The nested construction makes such sliding unlikely to matter. The visibility barrier is nondifferentiable at two kinds of point, the vertices of $\partial\mathcal{F}$ and the ties between features. The vertices lie on the boundary, at $h_i = 0$, which the observed agent approaches only when \eqref{eq:relaxed_vis} cannot be met, since that constraint drives $\mathbf{p}_{i-1}$ to a depth of at least $\epsilon\bar{d}$. The ties mostly lie deep inside the FoV, where $h_i$ is large and the strict constraint is unlikely to be active, so the choice of gradient there does not affect the command. A tie matters only near a vertex, where it coincides with a small $h_i$, and that region is again the one \eqref{eq:relaxed_vis} keeps the agent away from.

\section{Nested Field-of-View Barriers}
\label{sec:nested_fovs}

Each pair $i \in \mathcal{E}$ carries two visibility constraints: a strict one on the full FoV, split between the two agents of the pair, and a relaxed one on the contracted FoV, implemented by the observer alone.

\subsection{The Strict Distributed Constraint}
\label{sec:strict}

The barrier $h_i$ of \eqref{eq:h_full} couples two agents, so the CBF condition \eqref{eq:cbf_condition} for pair $i$,
\begin{equation}
    \frac{\partial h_i}{\partial \mathbf{x}_{i-1}} \mathbf{u}_{i-1} + \frac{\partial h_i}{\partial \mathbf{x}_{i}} \mathbf{u}_{i} \ge -\alpha_v h_i(\mathbf{x}),
    \label{eq:joint_vis}
\end{equation}
is a single inequality in two control inputs that no agent can enforce alone. We impose instead the two halves
\begin{equation}
\begin{aligned}
    \frac{\partial h_i}{\partial \mathbf{x}_{i-1}}(\mathbf{x})\, \mathbf{u}_{i-1} &\ge -\frac{\alpha_v}{2} h_i(\mathbf{x}), \\
    \frac{\partial h_i}{\partial \mathbf{x}_{i}}(\mathbf{x})\, \mathbf{u}_{i} &\ge -\frac{\alpha_v}{2} h_i(\mathbf{x}),
\end{aligned}
    \label{eq:split_vis}
\end{equation}
the first in the program of the observed agent $i-1$, the second in that of the observer $i$. Each is enforceable locally, each admits $\mathbf{u} = \mathbf{0}$ whenever $h_i \ge 0$, and their sum returns \eqref{eq:joint_vis} exactly. No slack appears in \eqref{eq:split_vis}. This is the strict constraint that carries the guarantee.

\subsection{The Relaxed Constraint on the Contracted FoV}
\label{sec:relaxed}

Enforcing \eqref{eq:split_vis} alone stalls the chain. The observer's half admits $\mathbf{u}_i = \mathbf{0}$ for every $h_i \ge 0$, and with no reference that is its minimizer, so the observer never moves. The observed agent's half caps its speed away from the observer at $\frac{\alpha_v}{2} h_i$, which vanishes at the boundary, so the leader decelerates toward the edge of the FoV of a stationary follower and stops there, connected but unable to advance. The contracted barrier $g_i$ of \eqref{eq:h_contracted} supplies the missing drive,
\begin{equation}
    \frac{\partial h_i}{\partial \mathbf{x}_{i}}(\mathbf{x})\, \mathbf{u}_{i} \ge -\alpha_c\, g_i(\mathbf{x}) - \delta_i, \qquad \delta_i \ge 0,
    \label{eq:relaxed_vis}
\end{equation}
where the gradient is that of $h_i$ by \eqref{eq:h_contracted}, so the constraint adds no geometric computation, and $\delta_i$ is a slack penalized in the cost. Whenever the observed agent leaves the contracted FoV, $g_i < 0$ and \eqref{eq:relaxed_vis} demands strictly positive progress from the observer, which moves inward while the strict constraint still has a margin of $\epsilon \bar{d}$. Where the environment makes that unattainable, $\delta_i$ absorbs the deficit and the pair falls back on \eqref{eq:split_vis}. The constraint is imposed on the observer alone, so it never restrains the teleoperated agent.

\subsection{Responsibility-Weighted Distributed Safety Constraints}
\label{sec:safety_split}

The inter-agent barrier $h_{a,ij}$ in \eqref{eq:h_safety} couples two agents in the same way, and is split by a pair of responsibility weights $\rho_{ij}, \rho_{ji} \in (0,1)$ with $\rho_{ij} + \rho_{ji} = 1$,
\begin{equation}
    \frac{\partial h_{a,ij}}{\partial \mathbf{x}_{i}}(\mathbf{x})\, \mathbf{u}_{i} \ge -\rho_{ij}\, \alpha_a\, h_{a,ij}(\mathbf{x}),
    \label{eq:split_agent}
\end{equation}
whose sum over the pair again returns the centralized condition. A smaller weight tightens the constraint, so setting $\rho_{ij} < \rho_{ji}$ for $i > j$ charges the agent further down the chain with the larger share of the avoidance effort and improves the teleoperated maneuverability. The obstacle barrier $h_{o,ik}$ depends on $\mathbf{x}_i$ alone and needs no splitting,
\begin{equation}
    \frac{\partial h_{o,ik}}{\partial \mathbf{x}_i}(\mathbf{x})\, \mathbf{u}_i \ge -\alpha_o\, h_{o,ik}(\mathbf{x}).
    \label{eq:obstacle_constraint}
\end{equation}

\subsection{The Individual Quadratic Program}
\label{sec:qp_agent}

Collecting \eqref{eq:split_vis}--\eqref{eq:obstacle_constraint}, each agent $i \in \{2, \dots, N-1\}$ solves, at state $\mathbf{x}$,
\begin{equation}
\begin{aligned}
    \min_{\mathbf{u}_i \in \mathcal{U},\, \delta_i} \ \ & (\mathbf{u}_i - \mathbf{r}_i)^\top \mathbf{P} (\mathbf{u}_i - \mathbf{r}_i) + w\, \delta_i^2 \\
    \text{s.t.} \ \ & \frac{\partial h_{i+1}}{\partial \mathbf{x}_{i}} \mathbf{u}_{i} \ge -\frac{\alpha_v}{2} h_{i+1}(\mathbf{x}), \\
    & \frac{\partial h_{i}}{\partial \mathbf{x}_{i}} \mathbf{u}_{i} \ge -\frac{\alpha_v}{2} h_{i}(\mathbf{x}), \\
    & \frac{\partial h_{i}}{\partial \mathbf{x}_{i}} \mathbf{u}_{i} \ge -\alpha_c\, g_{i}(\mathbf{x}) - \delta_i, \\
    & \frac{\partial h_{o,ik}}{\partial \mathbf{x}_i} \mathbf{u}_{i} \ge -\alpha_o h_{o,ik}(\mathbf{x}), \quad \forall\, \mathcal{O}_k \subset \mathcal{O}, \\
    & \frac{\partial h_{a,ij}}{\partial \mathbf{x}_{i}} \mathbf{u}_{i} \ge -\rho_{ij} \alpha_a h_{a,ij}(\mathbf{x}), \quad \forall\, j \neq i, \\
    & \delta_i \ge 0,
\end{aligned}
\label{eq:qp_agent}
\end{equation}
with $\mathbf{P} \succ 0$, $w > 0$, and $\mathbf{r}_i = \mathbf{0}$ for $i > 1$. Setting $\mathbf{r}_i = \mathbf{0}$ for $i > 1$ reflects the scenario studied here, not a requirement of the formulation. Feasibility and forward invariance depend only on the constraints of \eqref{eq:qp_agent}, never on its cost. A relay can therefore be given an objective of its own, encoded as a nonzero reference $\mathbf{r}_i$, and the program filters that objective rather than replacing it. The relaxed constraint gives way to the reference through the slack $\delta_i$, to a degree set by the weight $w$, while the strict constraints do not give way at all. Three adaptations close the chain:
\begin{itemize}
    \item agent $1$ observes no one, so both constraints indexed by $h_1$ are absent and $\mathbf{r}_1(t)$ is the operator command;
    \item the observer of pair $N$ is the static base, which solves no program. Agent $N-1$ therefore carries that pair undivided, with $-\alpha_v h_N$ in place of the first constraint of \eqref{eq:qp_agent}, and additionally carries its relaxed constraint, $\frac{\partial h_{N}}{\partial \mathbf{x}_{N-1}} \mathbf{u}_{N-1} \ge -\alpha_c g_{N} - \delta_N$, so that the link to the base is drawn inward like every other;
    \item agent $N$ is static, $\mathbf{u}_N \equiv \mathbf{0}$.
\end{itemize}

\section{Feasibility and Forward Invariance}
\label{sec:method2}

The method in Section~\ref{sec:nested_fovs} is verified in two steps. Lemma~\ref{lem:feasibility} shows that no program in \eqref{eq:qp_agent} can become infeasible on $\mathcal{S}$, so the control law is everywhere defined; Theorem~\ref{thm:invariance} shows that the distributed control inputs jointly satisfy the coupled conditions, so $\mathcal{S}$ is never left.

\begin{lemma}
\label{lem:feasibility}
Let Assumption~\ref{as:info} hold. For every $\mathbf{x} \in \mathcal{S}$ and every agent $i$, the point $(\mathbf{u}_i, \delta_i) = \big(\mathbf{0},\, [-\alpha_c g_i(\mathbf{x})]_+\big)$ is feasible for \eqref{eq:qp_agent}, where $[\,\cdot\,]_+ = \max(\cdot, 0)$. Consequently \eqref{eq:qp_agent} attains a unique minimizer at every $\mathbf{x} \in \mathcal{S}$.
\end{lemma}

\begin{proof}
Fix $\mathbf{x} \in \mathcal{S}$ and an agent $i$. Setting $\mathbf{u}_i = \mathbf{0}$ annihilates the left-hand side of every constraint of \eqref{eq:qp_agent}, which then reduce to sign conditions on the right-hand sides. The two strict visibility constraints require $0 \ge -\frac{\alpha_v}{2} h_{i+1}$ and $0 \ge -\frac{\alpha_v}{2} h_{i}$, satisfied because $\alpha_v > 0$ and $h_i, h_{i+1} \ge 0$ on $\mathcal{S}_v$; likewise the obstacle and inter-agent constraints hold because $h_{o,ik} \ge 0$ on $\mathcal{S}_o$, $h_{a,ij} \ge 0$ on $\mathcal{S}_a$, and $\alpha_o, \alpha_a, \rho_{ij} > 0$. The undivided variant carried by agent $N-1$ differs only in the constant $\alpha_v$ and is covered by the same argument. The relaxed constraint requires $0 \ge -\alpha_c g_i - \delta_i$, which the choice $\delta_i = [-\alpha_c g_i]_+ \ge 0$ satisfies for either sign of $g_i$, since $g_i$ is not constrained on $\mathcal{S}$.

The feasible set is thus nonempty, and it is a closed convex polyhedron as a finite intersection of half-spaces with $\mathcal{U}$ convex. The cost is strictly convex and coercive in $(\mathbf{u}_i, \delta_i)$ because $\mathbf{P} \succ 0$ and $w > 0$, so it attains a minimum on that set, and the minimizer is unique.
\end{proof}

With the control law defined at every point of $\mathcal{S}$, it remains to show that the inputs the agents compute independently keep the team there.

\begin{theorem}
\label{thm:invariance}
Let Assumptions~\ref{as:init}--\ref{as:transversal} hold and let each agent apply the minimizer of \eqref{eq:qp_agent}. Then the closed-loop team trajectory satisfies $\mathbf{x}(t) \in \mathcal{S}$ for all $t \ge 0$. In particular $\mathbf{p}_{i-1}(t) \in \mathcal{F}(\mathbf{x}_i(t))$ for every pair $i \in \mathcal{E}$ and every $t \ge 0$, so the leader is connected to the base by $N-1$ unoccluded, in-range links at all times.
\end{theorem}

\begin{proof}
By Lemma~\ref{lem:feasibility} the control law is defined at every $\mathbf{x} \in \mathcal{S}$, and by Assumption~\ref{as:lipschitz} it is locally Lipschitz there, so from $\mathbf{x}(0) \in \mathcal{S}$ the closed loop admits a unique solution on a maximal interval $[0,T)$ along which $\mathbf{x}(t) \in \mathcal{S}$. We verify that each barrier satisfies the CBF condition \eqref{eq:cbf_condition} on that interval.

\emph{\textbf{Obstacles.}} $h_{o,ik}$ depends on $\mathbf{x}_i$ alone, so $\dot{h}_{o,ik} = \frac{\partial h_{o,ik}}{\partial \mathbf{x}_i} \mathbf{u}_i^*$, and the fourth constraint of \eqref{eq:qp_agent}, satisfied by $\mathbf{u}_i^*$, gives $\dot{h}_{o,ik} \ge -\alpha_o h_{o,ik}$.

\emph{\textbf{Inter-agent safety.}} $h_{a,ij}$ depends on $\mathbf{x}_i$ and $\mathbf{x}_j$ only, so
\[
    \dot{h}_{a,ij} = \frac{\partial h_{a,ij}}{\partial \mathbf{x}_i} \mathbf{u}_i^* + \frac{\partial h_{a,ij}}{\partial \mathbf{x}_j} \mathbf{u}_j^*.
\]
By Assumption~\ref{as:info} both programs are solved at the same $\mathbf{x}$, so both instances of \eqref{eq:split_agent} hold simultaneously and, added, give $\dot{h}_{a,ij} \ge -(\rho_{ij} + \rho_{ji}) \alpha_a h_{a,ij} = -\alpha_a h_{a,ij}$.

\emph{\textbf{Visibility ($2 \le i \le N-1$).}} The barrier $h_i$ of \eqref{eq:h_full} is a function of $\mathbf{p}_{i-1}$ and $\mathbf{x}_i$ only, hence $\partial h_i / \partial \mathbf{x}_j = \mathbf{0}$ for $j \notin \{i-1, i\}$ and
\[
    \dot{h}_i = \frac{\partial h_i}{\partial \mathbf{x}_{i-1}} \mathbf{u}_{i-1}^* + \frac{\partial h_i}{\partial \mathbf{x}_{i}} \mathbf{u}_{i}^*.
\]
The first term is bounded below by $-\frac{\alpha_v}{2} h_i$ because $\mathbf{u}_{i-1}^*$ is feasible for the program of agent $i-1$, which contains the first constraint of \eqref{eq:split_vis}; the second likewise through the program of agent $i$. Both are evaluated at the same $\mathbf{x}$ by Assumption~\ref{as:info}, so adding them yields $\dot{h}_i \ge -\alpha_v h_i$, which is \eqref{eq:joint_vis}.

\emph{\textbf{Visibility at the base ($i = N$).}} Agent $N$ is static, so $\frac{\partial h_N}{\partial \mathbf{x}_N} \mathbf{u}_N = 0$ and $\dot{h}_N = \frac{\partial h_N}{\partial \mathbf{x}_{N-1}} \mathbf{u}_{N-1}^*$. Agent $N-1$ carries the undivided constraint, giving $\dot{h}_N \ge -\alpha_v h_N$ directly.

Every barrier of \eqref{eq:team_safe_set} therefore satisfies \eqref{eq:cbf_condition} along the solution. For the visibility barriers this holds at every $t$ at which the barrier is differentiable at $\mathbf{x}(t)$, hence at almost every $t \in [0,T)$ by Assumption~\ref{as:transversal}. Each barrier composed with $\mathbf{x}(\cdot)$ is absolutely continuous, being a locally Lipschitz function (Assumption~\ref{as:barrier}) of an absolutely continuous argument, and satisfies its differential inequality at almost every $t$, which is all the comparison argument behind Lemma~\ref{lem:invariance} requires \cite[Lem.~2]{glotfelter2017nonsmooth}. Each superlevel set is therefore forward invariant on $[0,T)$, and so is their intersection $\mathcal{S}$. Hence every barrier remains nonnegative on $[0,T)$ and, by continuity of $\mathbf{x}(\cdot)$ and of the barriers, at $t = T$ as well, so $\mathbf{x}(T) \in \mathcal{S}$. Lemma~\ref{lem:feasibility} and Assumption~\ref{as:lipschitz} then apply at $\mathbf{x}(T)$ and the solution extends beyond $T$, contradicting maximality unless $T = \infty$. Finally $\mathbf{x}(t) \in \mathcal{S}_v$ reads $d(\mathbf{p}_{i-1}, \mathcal{F}(\mathbf{x}_i)) \le 0$, which is $\mathbf{p}_{i-1} \in \mathcal{F}(\mathbf{x}_i)$ by Definition~\ref{def:sdf}.
\end{proof}

\begin{proposition}
\label{prop:neutral}
The set $\mathcal{S}$ and the conclusion of Theorem~\ref{thm:invariance} are independent of $\epsilon$, $\alpha_c$, and $w$.
\end{proposition}

\begin{proof}
The barriers $g_i$ do not enter \eqref{eq:team_safe_set}, and the proof of Theorem~\ref{thm:invariance} uses \eqref{eq:relaxed_vis} only through the value of $\mathbf{u}_i^*$, which exists for every $\epsilon$, $\alpha_c$, $w$ by Lemma~\ref{lem:feasibility}.
\end{proof}

The guarantee thus follows from the strict constraints alone, and $\epsilon$, $\alpha_c$ may be tuned freely, even to values that make the relaxed constraint frequently unattainable, without placing connectivity at risk.

\begin{remark}
\label{rem:workspace}
A hard connectivity guarantee bounds the workspace. Agent $1$ cannot exceed $N-1$ chained FoV ranges from the static base, and at that limit it is braked by the first constraint of \eqref{eq:split_vis}. This is intrinsic to any finite relay budget. The contracted constraint changes only when the operator reaches that limit. Since \eqref{eq:relaxed_vis} acts $\epsilon\bar{d}$ ahead of the strict boundary on each link, the followers pull their predecessors back inward while the strict constraint is still inactive. The leader therefore comes to a full stop only when the chain is fully extended, not whenever a single link momentarily approaches its boundary.
\end{remark}


\section{Simulation Setup}
\label{sec:setup}

Both systems realize the planar model of Section~\ref{sec:problem}, with the operator replaced by a reference law that drives agent~$1$ from the base to a goal. The nested construction is evaluated against each of its two visibility constraints enforced alone, in two workspaces and on two simulated systems, with Table~\ref{tab:ablation} carrying every number quoted below.

\subsection{Simulated Systems}

Both simulated systems run the same ROS~2 stack and differ only in what executes the velocity command. Each of the $N = 4$ agents solves its program \eqref{eq:qp_agent} with \textit{Gurobi} and publishes on the same topics at a control period $T_s = 0.2$\,s. Under the kinematic system, those commands are integrated by nodes implementing \eqref{eq:integrator}, so the system matches the model of Section~\ref{sec:preliminaries} exactly. It tests the construction only against the two effects the continuous-time analysis leaves out, sampling of the control law and ray casting of the field of view. Under the \emph{SIL} system the quadrotors fly at a common, fixed altitude, and the commands are tracked by \textit{Crazyflie} firmware. The firmware runs in \textit{CrazySim}~\cite{LlanesICRA2024}, interfaced through \textit{Crazyswarm2}~\cite{crazyswarm}, against quadrotor dynamics simulated in Gazebo, and its inner control loop is not modeled by \eqref{eq:integrator}. Team size and period are what the simulator sustains, four agents holding $5$\,Hz under Gazebo's load, and the executed rate never fell below $4.9$\,Hz, so no measurement below comes from a loop that missed its period. The kinematic system satisfies the assumption of Section~\ref{sec:agent_model} that the velocity command is tracked fast enough, and CrazySim adds the quadrotor dynamics and firmware that the kinematic system abstracts away.

\subsection{Workspaces}

\begin{figure}[t]
    \centering
    \vspace*{8pt}
    \begin{subfigure}{0.49\linewidth}\centering
        \includegraphics[width=0.9\linewidth]{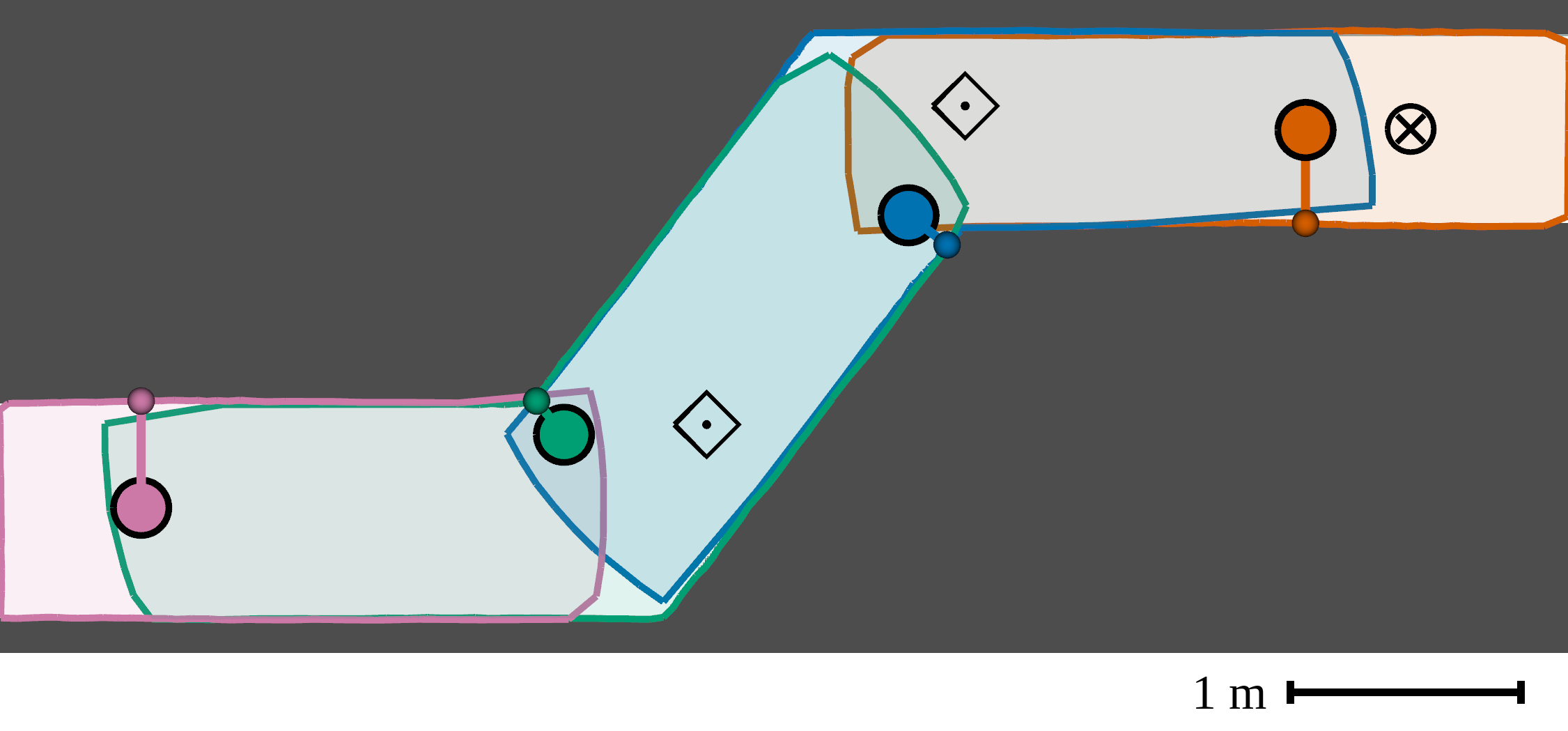}
        \caption{Tunnel, omnidirectional.}
        \label{fig:snap_tunnel}
    \end{subfigure}
    \hfill
    \begin{subfigure}{0.49\linewidth}\centering
        \includegraphics[width=0.9\linewidth]{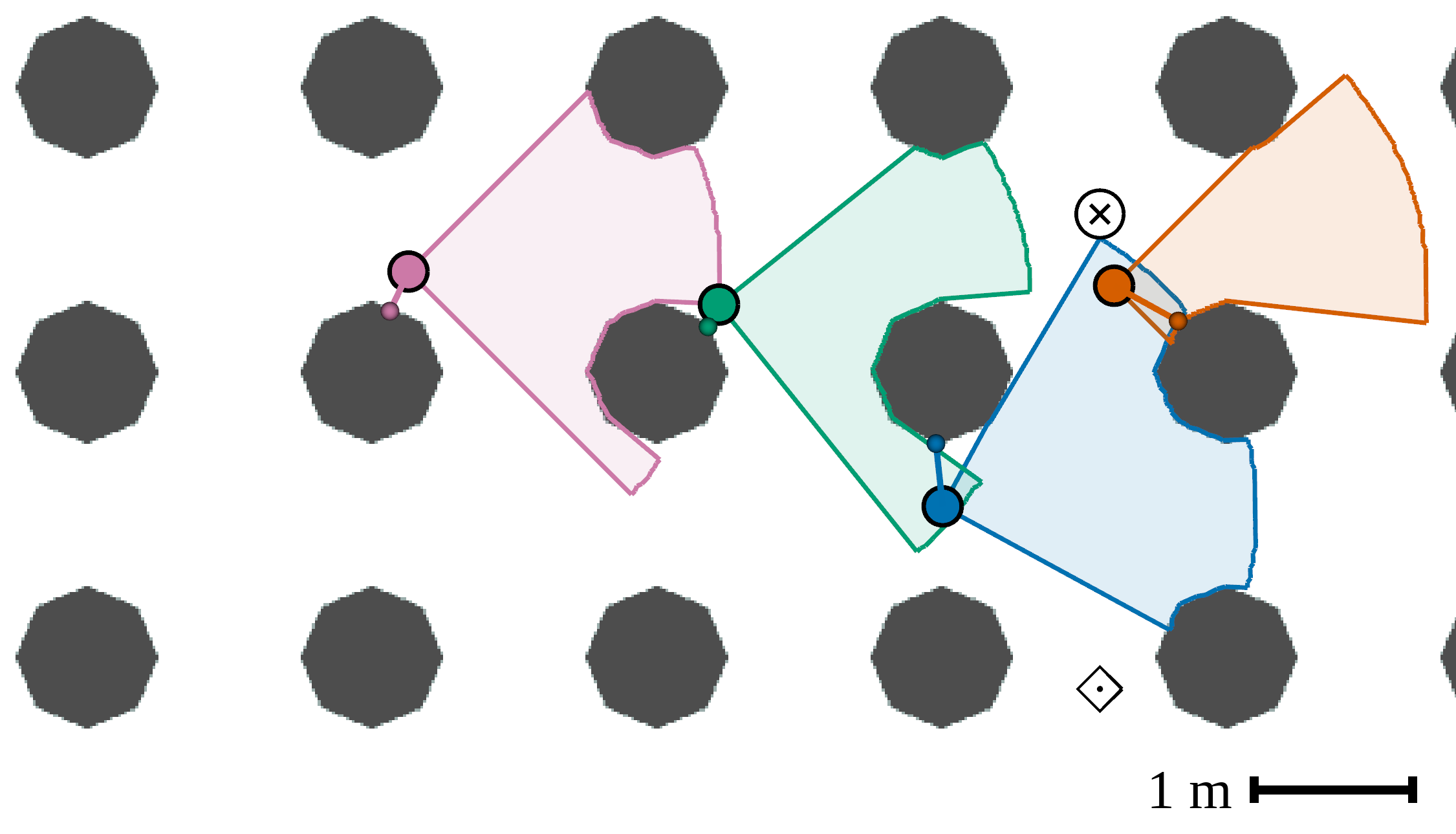}
        \caption{Clutter, $\pi/2$ aperture.}
        \label{fig:snap_cluttered}
    \end{subfigure}
    \caption{\textsc{Nested} at the instant the leader reaches the goal, with the waypoints marked $\diamond$ and the goal $\otimes$, each observer's occluded field of view, and the segment to its nearest obstacle.}
    \label{fig:snapshots}
\end{figure}

Both workspaces are represented as occupancy grids. The tunnel is $6.8 \times 3.77$\,m and the team is driven down it to a goal $4.19$\,m from the base, within the $(N-1)\,d_{\max} = 6.0$\,m relay budget of Remark~\ref{rem:workspace}, with the corridor's turns consuming the difference (Fig.~\ref{fig:snap_tunnel}). The cluttered field holds free-standing obstacles in rows, and the route crosses one of them diagonally before turning up the far side to a goal $2.97$\,m from the base (Fig.~\ref{fig:snap_cluttered}). The occluded FoV is ray cast over the grid with $100$ rays and its signed distance and gradient taken numerically from that polygon. Sensing has range $d_{\max} = 2.0$\,m and is omnidirectional in the tunnel, where $\theta_i$ is therefore left unconstrained, and restricted to a $\pi/2$ aperture in clutter, where it rotates with its observer.

\subsection{Operator and Parameters}

To make the comparison reproducible the operator is replaced by a reference law that drives agent $1$ at $v_{\mathrm{ref}} = 0.3$\,m/s toward a target advancing along the waypoints of Fig.~\ref{fig:snapshots}, the last of which is the goal. Parameters are $\mathbf{P} = \mathbf{I}$, $w = 2.0$, $\alpha_v = 0.8$, $\alpha_c = \alpha_o = \alpha_a = 1$, $R = 0.075$\,m inflating a $0.05$\,m vehicle radius, $d_o = 0.15$\,m measured from the agent's center, $\rho_{ij} = 0.5$, and a contraction $\epsilon\bar{d} = 0.8$\,m, one soft margin per link.

\subsection{Metrics}

Over each run we record $h_{\min} = \min_t \min_{i} h_i$, the closest approach to the boundary of the connected set, which Theorem~\ref{thm:invariance} requires nonnegative; the accumulated excursion $V_h = \sum_i \int_0^T [-h_i]_+ dt$ outside the full FoV, which is loss of connectivity; $\Delta T$, the arrival time in excess of a leader flying alone under its obstacle constraints only; the follower effort $E_f$, the commanded speed of the relays integrated over the run; $J = \max_t \max_i |h_i(t) - h_i(t - T_s)|$, the largest change of a visibility barrier across one step, which Section~\ref{sec:safe_set} bounds by $d_{\max}/d_o$ times the distance an agent covers in that step, here $0.80$\,m; and the clearances $d_a^{\min}$, $d_o^{\min}$. Each configuration is run $10$ times per system. The barrier metrics are reported as the worst value over the ten runs rather than the mean, since a single violation falsifies a guarantee and an average would conceal it.

\section{Results}
\label{sec:results}

Three controllers are compared under identical conditions. \textsc{Strict} imposes the split constraint \eqref{eq:split_vis} on both agents of each pair and nothing else on visibility. It is the barrier of \cite{koga2023hide, zhou2025control} without the slack an exogenous evader makes necessary, and a degenerate limit rather than a competing design. \textsc{Relaxed} gives both halves a penalized slack and imposes them on the contracted barrier $g_i$, so connectivity is a soft objective. Its slack puts it closer than \textsc{Strict} to the form of \cite{koga2023hide, zhou2025control}. \textsc{Nested} is the controller of Section~\ref{sec:nested_fovs}. \textsc{Relaxed} and \textsc{Nested} share $\epsilon$, so they differ only in whether the strict row is present. \textsc{Relaxed} carries one free parameter with no counterpart in \textsc{Nested}, the weight on its follower slack, and it was tuned in the baseline's favor in each workspace on the kinematic system, and carried unchanged to SIL, until its arrival came close to \textsc{Nested}'s; the comparison below is therefore at comparable arrival, with the baseline at its best setting rather than a convenient one.

\subsection{The Stall}

\textsc{Strict} stops the chain where it starts, timing out in all $10$ runs of both workspaces and both systems. A follower carrying only its half of \eqref{eq:split_vis} is inert while $h_i \ge 0$, for the reason given in Section~\ref{sec:relaxed}, and the leader's half then caps its speed against a stationary follower. The leader parks on the boundary and stays there, recording $h_{\min} = 6.6 \times 10^{-12}$\,m in every kinematic tunnel run, with $E_f$ below $0.5$\,m. This is the argument of Section~\ref{sec:introduction} made quantitative. The barrier certifies a following motion that nothing in the followers' programs produces.

\subsection{The Guarantee Without a Penalty}

On the kinematic system, which realizes the model of Theorem~\ref{thm:invariance}, the guarantee holds exactly in the tunnel, and to within $18$\,mm in clutter. \textsc{Nested} keeps $h_{\min} = +1.2 \times 10^{-3}$\,m with $V_h = 0$ in all ten kinematic tunnel runs, while \textsc{Relaxed} falls to $-0.228$\,m and accumulates $V_h = 0.366$\,m\,s. The slack that keeps the fleet mobile is the same slack that surrenders the link.

Holding the link costs no more than losing it. The two reach the goal at comparable times, $\Delta T = 12.1$\,s against its $13.2$\,s in the tunnel and $10.2$\,s each in clutter. The difference lies entirely in the relays' motion, $E_f = 5.95$\,m against $7.42$\,m in the tunnel and $5.99$\,m against $8.04$\,m in clutter, a fifth and a quarter less motion for an intact chain. Nesting therefore does not trade mobility for connectivity. The leader is nonetheless slowed for most of the run, the reference law never easing off, so that the strict row becomes the team's speed limit.

\subsection{Software-in-the-Loop}

On the SIL system, \textsc{Nested} again leads both baselines. \textsc{Relaxed} loses the link $2.9$ times deeper than \textsc{Nested} in the tunnel and $5.2$ times deeper in clutter, and \textsc{Strict}, comparable in depth, accumulates nine and four times more excursion (Table~\ref{tab:ablation}). The remaining excursions match the tracking error of the firmware. Comparing each command with the logged state one period later, over the forty SIL runs, the achieved velocity differs from the commanded one by $7.1 \pm 6.3$\,cm/s and the agent from its setpoint by $2.25 \pm 1.69$\,cm, a disturbance in \eqref{eq:cbf_condition} of the size that accounts for them. The controllers differ in how they absorb it. \textsc{Strict} parks on $h_i = 0$, so every tracking error becomes an excursion, and \textsc{Relaxed} has no slack-free row to bound it. \textsc{Nested} holds the link off the boundary, so its excursions stay shallow and brief, occurring only where the chain is fully extended. Obstacle clearance holds on the kinematic system, $d_o^{\min} \ge 0.150$\,m, and in SIL falls short by at most $0.028$\,m, the same setpoint error, with no collision between agents recorded.


\subsection{Regularity in Clutter}
\label{sec:cluttered}

The cluttered field exercises every moving feature of the FoV at once, aperture edges travelling with the observer and shadow edges sweeping about the corners that cast them. The barriers remain regular, the largest one-step change being $J = 0.213$\,m against the $0.80$\,m the lever of Section~\ref{sec:safe_set} allows. \textsc{Nested} leaves the connected set by at most $18$\,mm, under $1\%$ of the sensing range, against $0.397$\,m for \textsc{Relaxed}. This residual comes from discretization, in space by the $100$-beam ray cast, where a beam crossing an obstacle corner makes $h_i$ jump, and in time by the control period. It grows to $41$\,mm at a quarter of the beam density, and shrinks further only when both the beam gap and the distance covered per step are refined, as a sweep over beam count and control rate confirms.

\begin{table*}[t]
    \vspace*{10pt}
    \caption{Both workspaces, $10$ runs per cell, every entry kinematic\,/\,SIL. $h_{\min}$, $V_h$ and $J$ are the worst value over the ten runs, $\Delta T$ and $E_f$ their mean and standard deviation.}
    \label{tab:ablation}
    \centering
    \footnotesize
    \setlength{\tabcolsep}{3pt}
    \begin{tabular}{@{}lcccccc@{}}
        \toprule
        & \multicolumn{3}{c}{Tunnel} & \multicolumn{3}{c}{Clutter} \\
        \cmidrule(lr){2-4} \cmidrule(lr){5-7}
        & \textsc{Strict} & \textsc{Relaxed} & \textsc{Nested} & \textsc{Strict} & \textsc{Relaxed} & \textsc{Nested} \\
        \midrule
        $h_{\min}$ [m]     & $0.000/{-}0.127$ & ${-}0.228/{-}0.201$ & $0.001/{-}0.070$ & ${-}0.003/{-}0.073$ & ${-}0.397/{-}0.404$ & ${-}0.018/{-}0.078$ \\
        $V_h$ [m\,s]       & $0.000/0.618$    & $0.366/0.209$     & $0.000/0.071$    & $0.001/0.676$     & $2.277/2.014$     & $0.017/0.159$    \\
        $\Delta T$ [s]     & --  & $13.2{\pm}0.3 / 11.9{\pm}1.4$ & $12.1{\pm}0.3 / 11.7{\pm}1.0$ & -- & $10.2{\pm}0.5 / 14.4{\pm}2.6$ & $10.2{\pm}0.8 / 10.0{\pm}2.0$ \\
        $E_f$ [m]          & $0.00 / 0.41{\pm}0.12$ & $7.42{\pm}0.15 / 7.87{\pm}0.35$ & $5.95{\pm}0.08 / 5.64{\pm}0.20$ & $0.00 / 0.25{\pm}0.08$ & $8.04{\pm}0.20 / 11.68{\pm}0.41$ & $5.99{\pm}0.18 / 5.99{\pm}0.48$ \\
        $J$ [m]            & $0.036/0.113$   & $0.110/0.127$   & $0.110/0.111$   & $0.052/0.107$   & $0.061/0.213$   & $0.089/0.117$   \\
                \bottomrule
    \end{tabular}
\end{table*}

\section{Conclusions}
\label{sec:conclusions}

We presented a distributed control law that keeps a teleoperated chain connected by nesting two field-of-view barriers on each relay pair, a strict one split between the two agents of the pair, which carries the guarantee, and a relaxed one on a contracted field of view, imposed on the observer alone, which sets the followers into motion. Feasibility and forward invariance hold for every $\epsilon$, $\alpha_c$ and $w$, so the contraction may be tuned on the operator's behalf without placing connectivity at risk. In simulation the strict row alone never leaves the start and the relaxed row alone surrenders the link, while the nested pair arrives and, at comparable arrival time, holds the links far closer to the connected set than the relaxed baseline, with less follower effort.

Future work will extend Theorem~\ref{thm:invariance} to a bounded tracking error, letting the buffer $\epsilon\bar{d}$ absorb the inner-loop error observed in software-in-the-loop simulation, and will bring the sampling of the control law and of the field of view within the guarantee, removing the millimetre-scale residual observed in clutter.

\bibliographystyle{ieeetr}
\bibliography{bibliography}

\end{document}